\pdfoutput=1

\documentclass[11pt]{article}

\usepackage[]{main}

\usepackage{times}
\usepackage{latexsym}

\usepackage[T1]{fontenc}

\usepackage[utf8]{inputenc}

\usepackage{times}
\usepackage{epsfig}
\usepackage{graphicx}
\usepackage{amsmath}
\usepackage{amssymb}
\usepackage{enumitem}
\usepackage{tabularx}
\usepackage{multirow}
\usepackage{epsfig}

\graphicspath{ {./images/} }
\usepackage[rightcaption]{sidecap}
\usepackage[export]{adjustbox}
\usepackage{wrapfig}

\usepackage{microtype}

\usepackage{hyperref}

\title{Comprehensive Multi-Modal Interactions for Referring Image Segmentation}

\author{Kanishk Jain \and Vineet Gandhi \\ CVIT, KCIS, IIIT Hyderabad, India \\ \texttt{kanishk5991@gmail.com},   \texttt{vgandhi@iiit.ac.in}}

\begin{document}
\maketitle
\begin{abstract}
We investigate Referring Image Segmentation (RIS), which outputs a segmentation map corresponding to the natural language description. Addressing RIS efficiently requires considering the interactions happening \emph{across} visual and linguistic modalities and the interactions \emph{within} each modality. Existing methods are limited because they either compute different forms of interactions \emph{sequentially} (leading to error propagation) or \emph{ignore} intramodal interactions. We address this limitation by performing all three interactions \emph{simultaneously} through a Synchronous Multi-Modal Fusion Module (SFM). Moreover, to produce refined segmentation masks, we propose a novel Hierarchical Cross-Modal Aggregation Module (HCAM), where linguistic features facilitate the exchange of contextual information across the visual hierarchy. We present thorough ablation studies and validate our approach's performance on four benchmark datasets, showing considerable performance gains over the existing state-of-the-art (SOTA) methods.
\end{abstract}

\section{Introduction}

Traditional computer vision tasks like detection and segmentation have dealt with a pre-defined set of categories, limiting their scalability and practicality. Substituting the pre-defined categories with natural language expressions (NLE) is a logical extension to counteract the above problems. Indeed, this is how humans interact with objects in their environment; for example, the phrase ``the kid running after the butterfly" requires localizing only the child running after the butterfly and not the other kids. Formally, the task of localizing objects based on NLE is known as Visual Grounding. Existing works either approach the grounding problem by predicting a bounding box around the referred object or a segmentation mask corresponding to the referred object. We focus on the latter approach, as a segmentation mask can effectively pinpoint the exact location and capture the actual shape of the referred object. The task is formally known as Referring Image Segmentation (RIS).

\begin{figure}[t]
\begin{center}
  \includegraphics[width=\linewidth]{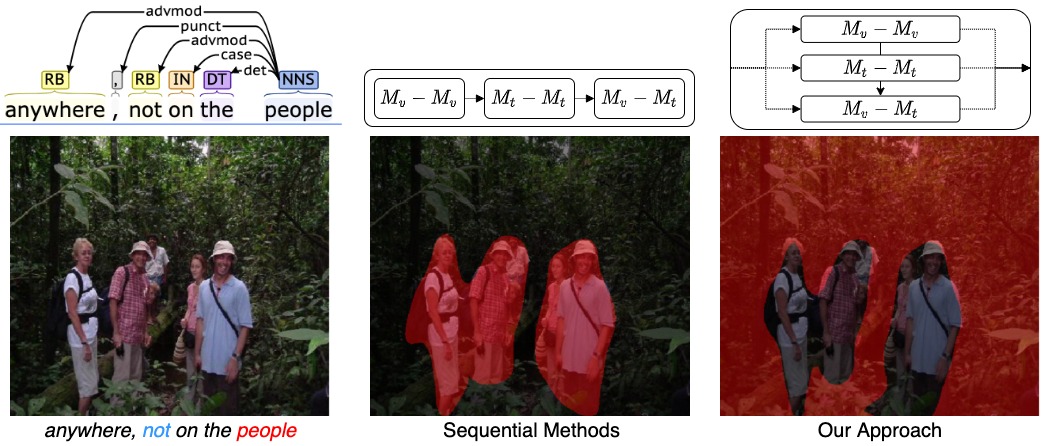}
\end{center}
  \caption{Unlike existing methods which model interactions in a sequential manner, we synchronously model the Intra-Modal and Inter-Modal interactions across visual and linguistic modalities. Here, $M_{v}$ and $M_{t}$ represent Visual and Linguistic Modalities, and \{-\} represents interactions between them.}
\label{fig:motivation}
\end{figure}

RIS requires understanding both visual and linguistic modalities at an individual level, specifically word-word and region-region interactions. Additionally, a mutual understanding of both modalities is required to identify the referred object from the linguistic expression and localize it in the image. For instance, to ground a sentence ``whatever is on the truck", it is necessary to understand the relationship between words as grounding just the individual words will not work. Similarly, region-to-region interactions in visual modality help group semantically similar regions, e.g., all regions belonging to the truck. Finally, to identify the referent regions, we need to transfer the distinctive information about the referent from the linguistic modality to the visual modality; this is taken care of by the cross-modal word-region interactions. The current SOTA methods~\cite{Yang_2021_CVPR, Feng_2021_CVPR, Huang_2020_CVPR, hui2020linguistic, Hu_2020_CVPR} take a modular approach, where these interactions happen in parts, sequentially.

Different methods differ in how they model these interactions. \cite{Huang_2020_CVPR} first perform a region-word alignment (cross-modal interaction). The second stage takes these alignments as input to select relevant image regions corresponding to the referent.  \cite{Yang_2021_CVPR} and \cite{hui2020linguistic} use the dependency tree structure of the referring expression for the reasoning stage instead. \cite{Hu_2020_CVPR} select a suitable combination of words for each region, followed by selecting the relevant regions corresponding to referent based on the affinities with other regions. The performance of the initial stages bounds these approaches. Furthermore, they ignore the crucial intra-modal interactions for RIS.


In this paper, we perform all three forms of interactions simultaneously. We propose a Synchronous Multi-Modal Fusion Module (SFM) which captures the inter-modal and intra-modal interactions between visual and linguistic modalities in a single step. Intra-modal interactions handle the cases for identifying the relevant set of words and semantically similar image regions. Inter-modal interactions transfer contextual information across modalities. Additionally, we propose a novel Hierarchical Cross-Modal Aggregation Module (HCAM) to exchange contextual information relevant to referent across visual hierarchies and refine the referred object's segmentation mask.

We motivate the benefits of simultaneous interactions over sequential in Figure \ref{fig:motivation} by presenting a failure case of the latter. For the given referring expression "anywhere, not on the people", sequential approaches fail to identify the correct word to be grounded, and the error gets propagated till the end. CMPC \cite{Huang_2020_CVPR} which predicts the referent word from the expression in the first stage, identifies "people" as the referent (middle image in Figure \ref{fig:motivation}) and completely misses "anywhere" which is the correct entity to ground. Similarly, \cite{Yang_2021_CVPR}, and \cite{hui2020linguistic}, which utilize dependency tree structure to govern their reasoning process, identify the referred entity "anywhere" as an adverb from the dependency tree. However, considering the expression in context with the image, the word "anywhere" should be perceived as a "pronoun". The proposed SFM module successfully addresses the aforementioned limitations. Overall, our work makes the following contributions:- 
\begin{enumerate} 
    \item We propose SFM to reason over regions, words, and region-word features in a synchronous manner, allowing each modality to focus on relevant semantic information to identify the referred object. \vspace{-0.8em}
    \item We propose a novel HCAM module, which routes hierarchical visual information through linguistic features to produce a refined segmentation mask. \vspace{-0.8em}
    \item We present thorough quantitative and qualitative experiments to demonstrate the efficacy of our approach and show notable performance gains on four RIS benchmarks. 
\end{enumerate}

\section{Related Work}



\begin{figure*}[ht]
\begin{center}
\includegraphics[width=\textwidth, inner]{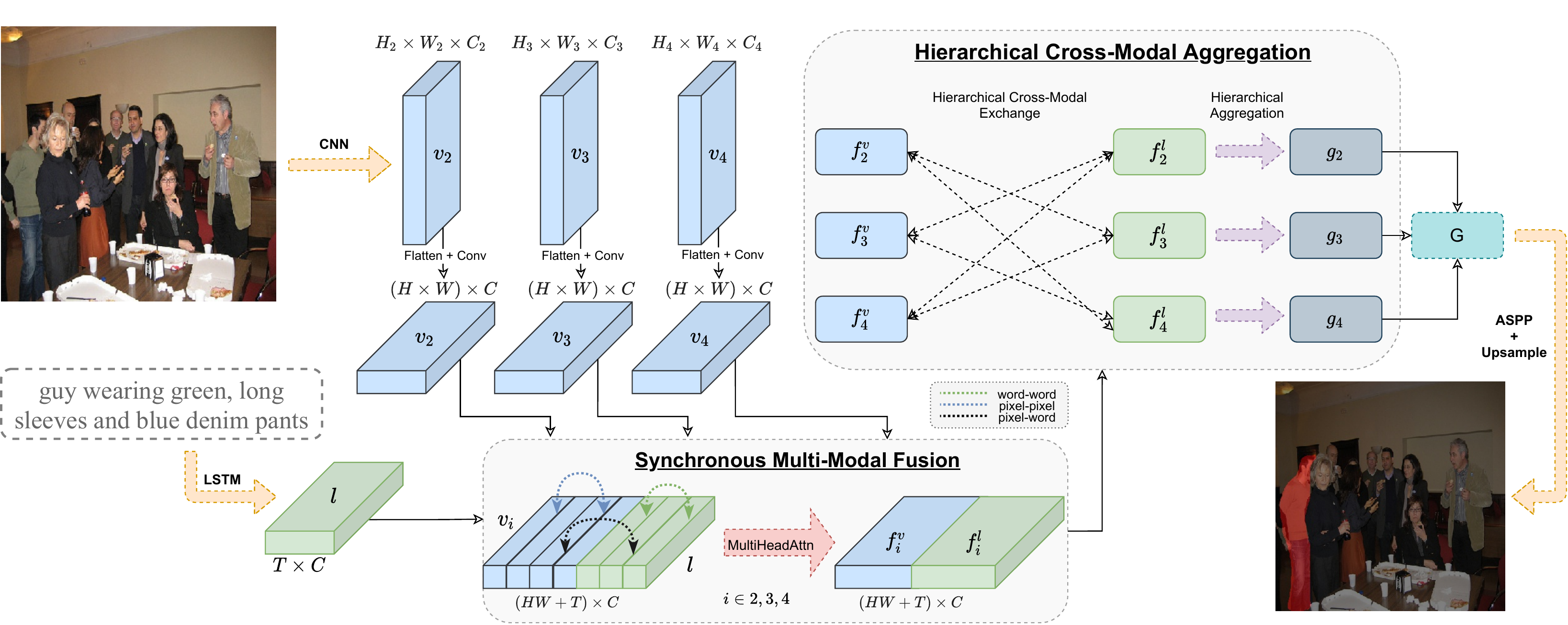}
\end{center}
   \caption{\centering{The proposed network architecture. Synchronous Multi-Modal Fusion captures pixel-pixel, word-word and pixel-word interaction. Hierarchical Cross-Modal Aggregation exchanges information across modalities and hierarchies to selectively aggregate context relevant to the referent.}}
\label{fig:arch}
\end{figure*}

\paragraph{Referring Expression Comprehension:} Localizing a bounding box/proposals based on an NLE is a task commonly referred to as Referring Expression Comprehension (REC). The majority of methods for REC learn a joint embedding space for visual and linguistic modalities and differ in how joint space is computed and how it is used. Earlier methods, \cite{hu2016natural, Rohrbach_2016, Plummer_2018_ECCV} used joint embedding space as a metric space to rank proposal features with linguistic features. Later methods like \cite{Yang2019CrossModalRI, Deng_2018_CVPR, Liu_Wan_Zhu_He_2020} utilized attention over the proposals to select the appropriate one. More Recent Methods like \cite{lu2019vilbert, 10.1007/978-3-030-58577-8_7} utilize transformer-based architecture to project multi-modal features to common semantic space. Specifically, they utilize a self-attention mechanism to align \emph{proposal-level features} with linguistic features. In our work, we utilize \emph{pixel-level image features} which are crucial for the task of RIS. Additionally, compared to  \cite{lu2019vilbert}, we \emph{explicitly} capture inter-modal and intra-modal interactions between visual and linguistic modalities. 

\paragraph{Referring Image Segmentation:} Bounding Box-based methods in REC are limited in their capabilities to capture the inherent shape of the referred object, which led to the proposal of the RIS task. It was first introduced in \cite{hu2016segmentation}, where they generate the referent's segmentation mask by directly concatenating visual features from CNN with tiled language features from LSTM. \cite{Li_2018_CVPR} generates refined segmentation masks by incorporating multi-scale semantic information from the image. Since each word in expression makes a different contribution in identifying the desired object, \cite{Shi_2018_ECCV} model visual context for each word separately using query attention. \cite{ye2019cross} uses a self-attention mechanism to capture long-range correlations between visual and textual modalities. Recent works \cite{Hu_2020_CVPR, Huang_2020_CVPR, hui2020linguistic} utilize cross-modal attention to model multi-modal context, \cite{hui2020linguistic, Yang_2021_CVPR} use dependency tree structure and \cite{Huang_2020_CVPR} use coarse labelling for each word in the expression for selective context modelling. Most of the existing works capture Inter and Intra modal interactions separately to model the context for referent. In this work, we \emph{concurrently} model the comprehensive interactions across visual and linguistic modalities.

 
 
\section{Method}
Given an image and a natural language referring expression, the goal is to predict a pixel-level segmentation mask corresponding to the referred entity described by the expression. The overall architecture of the network is illustrated in Figure \ref{fig:arch}. Visual features for the image are extracted using a CNN backbone, and linguistic features for the referring expression are extracted using a LSTM. A Synchronous Multi-Modal Fusion Module (SFM) simultaneously aligns visual regions with textual words and jointly reasons about both modalities to identify the multi-modal context relevant to the referent. SFM is applied to hierarchical visual features extracted from CNN backbone since hierarchical features are better suited for segmentation tasks~\cite{ye2019cross, Chen_2019_ICCV, Hu_2020_CVPR}. A novel Hierarchical Cross-Modal Aggregation module (HCAM) is applied to effectively fuse SFM's multi-level output and produce a refined segmentation mask for the referent. We describe the feature extraction process in the next section, and both SFM and HCAM modules are described in the subsequent sections.

\subsection{Feature Extraction}


Our network takes an image and a natural language expression as input. We extract hierarchical visual features for an image from a CNN backbone. Through pooling and convolution operations, all hierarchical visual features are transformed to the same spatial resolution and channel dimension. Final visual features for each level are of shape $\mathbb R^{C_v \times H \times W}$, with $H$, $W$ and $C_v$ being the height, width, and channel dimension of the visual features. Final visual features are denoted as $\{V_2, V_3, V_4\} $, corresponding to layers $2$, $3$ and $4$ of the CNN backbone. For ease of readability, we denote the visual features as $V$.  GloVe embeddings for each word in the referring expression are then passed as input to LSTM. The hidden feature of LSTM at $i^{th}$ time step $l_i \in \mathbb R^{C_l}$ , is used to denote the word feature for the $i^{th}$ word in the expression. The final linguistic feature of the expression is denoted as $ L = \{l_1, l_2, ..., l_T\} $, where $T$ is the number of words in the referring expression.

\subsection{Synchronous Multi-Modal Fusion}




In this section, we describe the Synchronous Multi-Modal Fusion Module (SFM). To successfully segment the referent, we need to identify the semantic information relevant to it in both the visual and linguistic modalities. We capture comprehensive intra-modal and inter-modal interactions explicitly in a synchronous manner, allowing us to jointly reason about visual and linguistic modalities while considering the contextual information from both.

Hierarchical visual features $V \in \mathbb R^{C_v \times H \times W} $ and linguistic word-level features $L \in \mathbb R^{C_l \times T} $ are passed as input to SFM, with $C_v = C_l = C$. We flatten the spatial dimensions of visual features and perform a lengthwise concatenation with linguistic feature, followed by layer normalization to get multi-modal feature $X$ of shape $\mathbb R^{C \times (HW + T)}$. We then add separate positional embedding $P_v$ and $P_l$ to visual $X_v \in \mathbb R^{C \times HW}$ and linguistic $X_l \in \mathbb R^{C \times T}$ part of $X$ to distinguish between visual and linguistic part. Finally, we apply multi-head attention over $X$ to capture the inter-modal and intra-modal interactions between visual and linguistic modalities. Specifically, pixel-pixel, word-word and word-pixel interactions are captured. Pixel-pixel and word-word interactions help in independently identifying semantically similar pixels and words in their respective modalities, pixel-word interaction helps in identifying corresponding pixels and words with similar contextual semantics across modalities.
\begin{equation} \label{eq1}
\begin{split}
X & = \text{LayerNorm}(V \odot L) \\
X & = X + (P_v \odot P_l) \\
F & = \text{MultiHead}(X)
\end{split}
\end{equation}
Here, $\odot$ is length-wise concatenation, $F$ is the final output of SFM module having same shape as $X$. We process all hierarchical visual features $\{V_2, V_3, V_4\}$ individually through SFM, resulting in hierarchical cross-modal output $\{F_2, F_3, F_4\}$.

\subsection{Hierarchical Cross-Modal Aggregation} 
\begin{figure}[t] 
    \centering
    \includegraphics[width=0.49\textwidth]{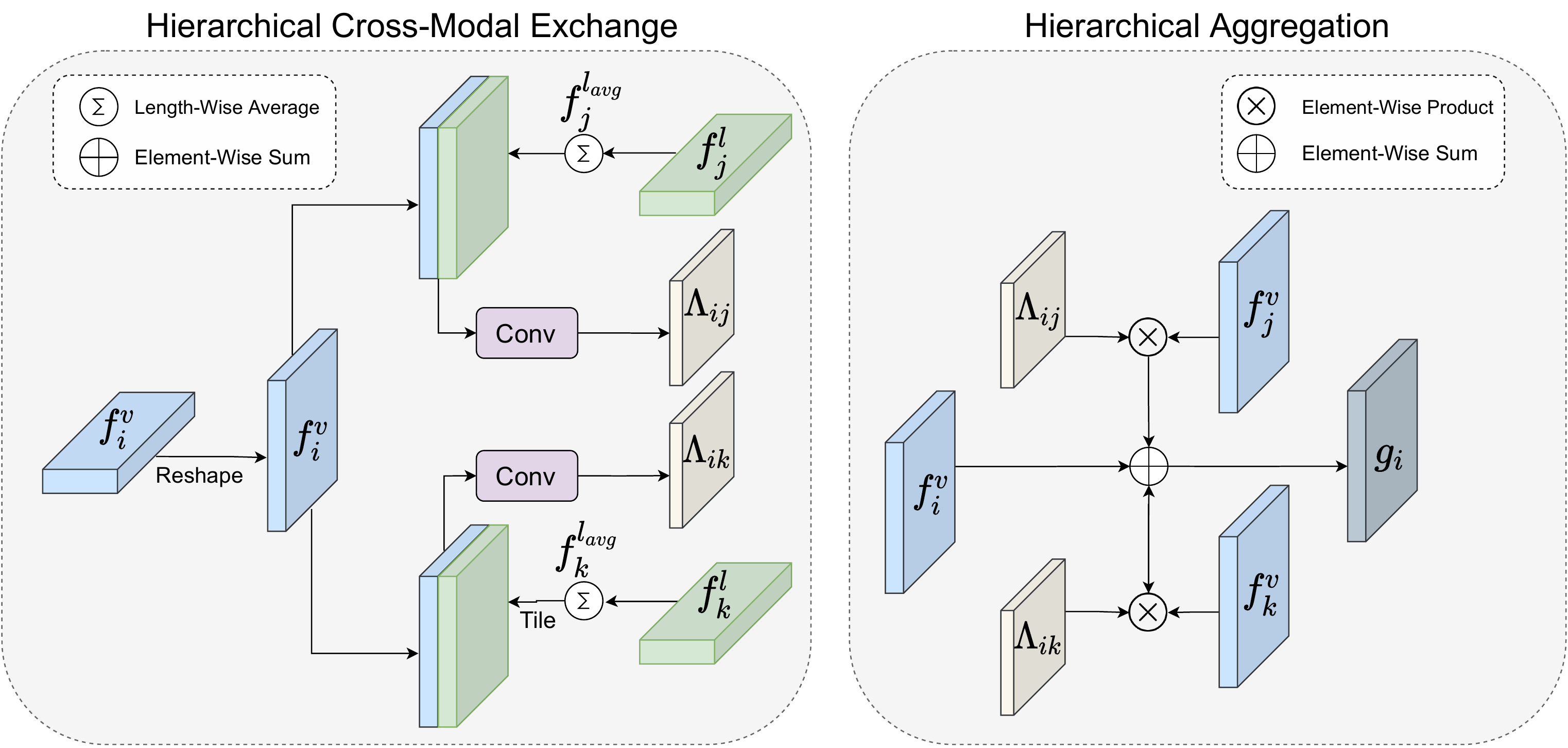}
    \caption{Our Novel Hierarchical Cross-Modal Aggregation Module consisting of Hierarchical Cross-Modal Exchange and Hierarchical Aggregation steps.}
    \label{fig:cmmlf}
\end{figure}

Hierarchical visual features of CNN capture different aspects of images. As a result, depending on the hierarchy, visual features can focus on different aspects of the linguistic expression. In order to predict a refined segmentation mask, different hierarchies should be in agreement regarding the image regions to focus on. Therefore, all visual hierarchical features should also focus on image regions corresponding to linguistic context from other hierarchies. This will ensure that all hierarchical features are focusing on common regions. We propose a novel Hierarchical Cross-Modal Aggregation (HCAM) module for this purpose. HCAM includes two key steps: (1) \textbf{Hierarchical Cross-Modal Exchange}, and (2) \textbf{Hierarchical Aggregation}. Both steps are illustrated in Figure~\ref{fig:cmmlf}.

\textbf{Hierarchical Cross-Modal Exchange:} During the HCME step, we calculate the affinity weights $\Lambda_{ij}$ between the $j^{th}$ layer's linguistic context $f_{j}^{l}$ and the spatial regions for $i^{th}$ layer's visual features $f_{i}^{v}$, where $f_{i}^{v}$ and $f_{i}^{l}$ are the visual and linguistic part of $i^{th}$ layer's output of SFM $F_i$.
\begin{equation} \label{eq2}
\begin{split}
\Lambda_{ij} = \sigma(Conv([f_{i}^{v}; f_{j}^{l_{avg}}]))
\end{split}
\end{equation}
Here $ \Lambda_{ij} \in \mathbb R^{C \times H \times W}$, $f_{j}^{l_{avg}} \in \mathbb R^{C}$ is the global linguistic context for $j^{th}$ layer and is computed as length-wise average of linguistic features $f_{j}^{l}$, $\sigma$ is the sigmoid function. Here, $f_{j}^{l_{avg}}$ act as a bridge to route linguistic context from $j^{th}$ layer to spatial regions of $i^{th}$ layer's visual hierarchy. Similarly, $ \Lambda_{ik}$ is computed with $i \neq j \neq k$, allowing for cross-modal exchange between all permutations of visual and linguistic hierarchical features.


\textbf{Hierarchical Aggregation:} After computing the affinity weights $\Lambda_{ij}$, we perform a layer-wise contextual aggregation. For each layer, visual context from other hierarchies is aggregated in the following way:
\begin{equation} \label{eq3}
\begin{split}
g_{i} = f_{i}^{v} + \sum_{j \neq i } \Lambda_{ij} \circ f_{j}^{v} \\ 
G = Conv3D([g_2; g_3; g_4])
\end{split}
\end{equation}
Here, $\circ$ is element-wise product and $[;]$ represents stacking features along length dimension, ie:- $\mathbb R^{3 \times C \times H \times W}$ dimensional feature. $g_{i} \in \mathbb R^{C \times H \times W}$ contains the relevant regions corresponding to the linguistic context from the other two hierarchies. Finally, we use 3D convolution to aggregate $g_{i}$'s to include the common regions corresponding to the linguistic context from all visual hierarchies. $G$ is the final multi-modal context for referent.


\subsection{Mask Generation}
Finally, $G$ is passed through Atrous Spatial Pyramid Pooling (ASPP) decoder \cite{chen2018encoderdecoder} and Up-sampling convolution to predict final segmentation mask $S$. Pixel-level binary cross-entropy loss is applied to predicted segmentation map $S$ and the ground truth segmentation mask $ Y $ to train the entire network end-to-end.

~\vspace{-1.5em}

\section{Experiments}

\begin{table*}[]
\small
\begin{center}
\begin{tabular}{c|ccc|ccc|c|c}
\hline
Method & \multicolumn{3}{c|}{UNC} & \multicolumn{3}{c|}{UNC+} & G-Ref & Referit \\ \cline{2-9} 
 & \multicolumn{1}{c|}{val} & \multicolumn{1}{c|}{testA} & testB & \multicolumn{1}{c|}{val} & \multicolumn{1}{c|}{testA} & testB & val & test \\ \hline
RRN~\cite{Li_2018_CVPR} & 55.33 & 57.26 & 53.95 & 39.75 & 42.15 & 36.11 & 36.45 & 63.63 \\
CMSA~\cite{ye2019cross} & 58.32 & 60.61 & 55.09 & 43.76 & 47.60 & 37.89 & 39.98 & 63.80 \\
STEP~\cite{Chen_2019_ICCV} & 60.04 & 63.46 & 57.97 & 48.19 & 52.33 & 40.41 & 46.40 & 64.13 \\
BRIN~\cite{Hu_2020_CVPR} & 61.35 & 63.37 & 59.57 & 48.57 & 52.87 & 42.13 & 48.04 & 63.46 \\
LSCM~\cite{hui2020linguistic} & 61.47 & 64.99 & 59.55 & 49.34 & 53.12 & 43.50 & 48.05 & 66.57 \\
CMPC~\cite{Huang_2020_CVPR} & 61.36 & 64.53 & 59.64 & 49.56 & 53.44 & 43.23 & 49.05 & 65.53 \\ 
BUSNet*~\cite{Yang_2021_CVPR} & 62.56 & 65.61 & 60.38 & 50.98 & 56.14 & 43.51 & \textcolor{red}{49.98} & - \\ 
EFN*~\cite{Feng_2021_CVPR} & 62.76 & 65.69 & 59.67 & 51.50 & 55.24 & 43.01 & \underline{51.93} & 66.70 \\ \hline
SHNet* $(320 \times 320)$ & \textcolor{blue}{63.98} & \textcolor{blue}{67.51} & \textcolor{blue}{60.48} & \textcolor{blue}{51.79} & \textcolor{blue}{56.49} & \textcolor{blue}{43.83} & 48.95 & \textcolor{blue}{68.38}  \\ 
SHNet* $(448 \times 448)$ & \textcolor{red}{65.32} & \textcolor{red}{68.56} & \textcolor{red}{62.04} & \textcolor{red}{52.75} & \textcolor{red}{58.46} & \textcolor{red}{44.12} & \textcolor{blue}{49.90} & \textcolor{red}{69.19}  \\ \hline

\end{tabular}
\end{center}
\caption{Comparison with State-Of-the-Arts on \textit{Overall IoU} metric, $*$ indicates results without using DenseCRF post processing. Best scores are shown in red and the second best are shown in blue. Our method uses DeepLabv3+ backbone for both resolutions.}
\label{table:overall_IoU}
\end{table*}


\subsection{Experimental Setup}
We conduct experiments on four Referring Image Segmentation datasets. \textbf{UNC} \cite{yu2016modeling} contains 19,994 images taken from MS-COCO~\cite{Lin2014MicrosoftCC} with 142,209 referring expressions corresponding to 50,000 objects. Referring Expressions for this dataset contain words indicating the location of the object. \textbf{UNC+} \cite{yu2016modeling} is also based on images from MS-COCO. It contains 19,992 images, with 141,564 referring expressions corresponding to 50,000 objects. In UNC+, the expression describes the object based on their appearance and context within the scene without using spatial words. \textbf{G-Ref} \cite{mao2016generation} is also curated using images from MS-COCO. It contains 26,711 images, with 104,560 referring expressions for 50,000 objects. G-Ref contains longer sentences with an average length of 8.4 words; compared to other datasets which have an average sentence length of less than 4 words. \textbf{Referit} \cite{KazemzadehOrdonezMattenBergEMNLP14} comprises of 19,894 images collected from IAPR TC-12 dataset. It includes 130,525 expressions for 96,654 objects. It contains unstructured regions (e.g., sky, mountains, and ground) as ground truth segmentations.

\subsection{Implementation details}

We experiment with two backbones, DeepLabv3+~\cite{chen2018encoderdecoder} and Resnet-101 for image feature extraction. Like previous works~\cite{ye2019cross, Chen_2019_ICCV, Hu_2020_CVPR}, DeepLabv3+ is pre-trained on Pascal VOC semantic segmentation task while Resnet-101 is pre-trained on Imagenet Classification task, and both backbone's parameters are fixed during training. For multi-level features, we extract features from the last three blocks of CNN backbone. We conduct experiments at two different image resolutions, $320 \times 320$ and $448 \times 448$ with $H=W=18$. We use GLoVe embeddings~\cite{Pennington14glove:global} pre-trained on Common Crawl 840B tokens to initialize word embedding for words in the expressions. The maximum number of words in the linguistic expression is set to 25. We use LSTM for extracting textual features. The network is trained using AdamW optimizer with batch size set to 50; the initial learning rate is set to $1.2e^{-4} $ and weight decay of $9e^{-5}$ is used. The initial learning rate is gradually decreased using polynomial decay with a power of 0.7. We train our network on each dataset separately. 


\textbf{Evaluation Metrics}: Following previous works~\cite{ye2019cross, Chen_2019_ICCV, Hu_2020_CVPR}, we evaluate the performance of our model using overall Intersection-over-Union (overall IoU) and Precision@$X$ as metrics. Overall IoU metric calculates the ratio of the intersection and the union computed between the predicted segmentation mask and the ground truth mask over all test samples. Precision@$X$ metric calculates the percentage of test samples having IoU greater than the threshold $X$, with $ X \in \{0.5, 0.6, 0.7, 0.8, 0.9\}$.

\begin{table*}[]
\small
\begin{center}
\begin{tabular}{r|c|c|c|c|c|c|c}
\hline
\multicolumn{1}{l|}{} &
  \multicolumn{1}{l|}{Method} &
  \multicolumn{1}{l|}{\textit{prec@0.5}} &
  \multicolumn{1}{l|}{prec@0.6} &
  \multicolumn{1}{l|}{prec@0.7} &
  \multicolumn{1}{l|}{prec@0.8} &
  \multicolumn{1}{l|}{prec@0.9} &
  \multicolumn{1}{l}{\textit{Overall IoU}} \\ \hline
1 & Baseline        & 61.47          & 54.01          & 43.74          & 27.47          & 7.21           & 54.70          \\
2 & Only HCAM      & 68.44          & 61.58          & 52.10          & 35.63          & 9.71           & 59.53          \\
3 & Only SFM        & 72.56          & 66.58          & 57.91          & 40.73          & 12.82          & 62.16          \\
4 & SFM+ConvLSTM    & 74.34          & 68.89          & 60.67          & 42.95          & 13.35          & 63.30          \\
5 & SFM+Conv3D      & 74.07          & 68.74          & 60.50          & 43.14          & 13.58          & 63.16          \\
6 & SHNet w/o Glove & 74.23          & 68.42          & 59.77          & 42.47          & 13.66          & 62.19          \\
7 & SHNet w/o P.E   & 74.0          & 68.36          & 59.71          & 43.15          & 13.36          & 63.07          \\
8 & SHNet           & \textbf{75.18} & \textbf{69.36} & \textbf{61.21} & \textbf{46.16} & \textbf{16.23} & \textbf{63.98} \\ \hline
\end{tabular}
\end{center}
\caption{Ablation Studies on Validation set of UNC, SHNet is the full architecture with both SFM and HCAM modules. The input image resolution is $320 \times 320$ in each case.}
\label{table:ablation_1}
\end{table*}

\subsection{Comparison with State of the Art}
We evaluate our method's performance on four benchmark datasets and present the results in Table \ref{table:overall_IoU}. Since three of the datasets are derived from MS-COCO and have significant overlap with each other, pre-training on MS-COCO can give misleading results and should be avoided. Hence, we only compare against methods for which the backbone is pre-trained on Pascal VOC. Unless specified, all the approaches in Table \ref{table:overall_IoU} are at $320 \times 320$ resolution. Our approach, SHNet (SFM+HCAM), achieves state-of-the-art performance on three datasets without post-processing. In contrast, most previous methods present results after post-processing through a Dense Conditional Random Field (Dense CRF).
The expressions in UNC+ avoid using positional words while referring to objects; instead, they are more descriptive about their attributes and relationships. Consistent performance gains on the UNC+ dataset at all splits showcases the effectiveness of utilizing comprehensive interactions simultaneously across visual and linguistic modalities. Similarly, our approach gains $1.68$\% over the next best performing method EFN~\cite{Feng_2021_CVPR} on the Referit dataset, reflecting its ability to ground unstructured regions (e.g., the sky, free space). We also achieve solid performance gains on the UNC dataset at both resolutions, indicating that our method can effectively utilize the positional words to localize the correct instance of an object from multiple ones. 
EFN~\cite{Feng_2021_CVPR} (underlined in Table~\ref{table:overall_IoU}) gives the best performance on G-Ref dataset; however, it is fine-tuned on the UNC pre-trained model. With similar fine-tuning, SHNet achieves 56.44\% overall IoU, surpassing EFN by a large margin. However, such an experimental setup is incorrect, as there is a significant overlap between G-Ref test and UNC training set. Hence, in Table \ref{table:overall_IoU} we report performance on a model trained on G-Ref from scratch. Performance of SHNet is marginally below BusNet on the G-Ref dataset. Feature maps in SHNet have a lower resolution of $18 \times 18$ compared to $40 \times 40$ resolution used by other methods and that possibly leads to a drop in performance on G-Ref, which has extremely small target objects. We could not train SHNet on higher resolution feature maps due to memory limits induced by multi-head attention (on RTX 2080Ti GPU); however, training on higher resolution input improves results. 

\begin{figure*}[t] 
    \centering
    \includegraphics[width=0.95\textwidth]{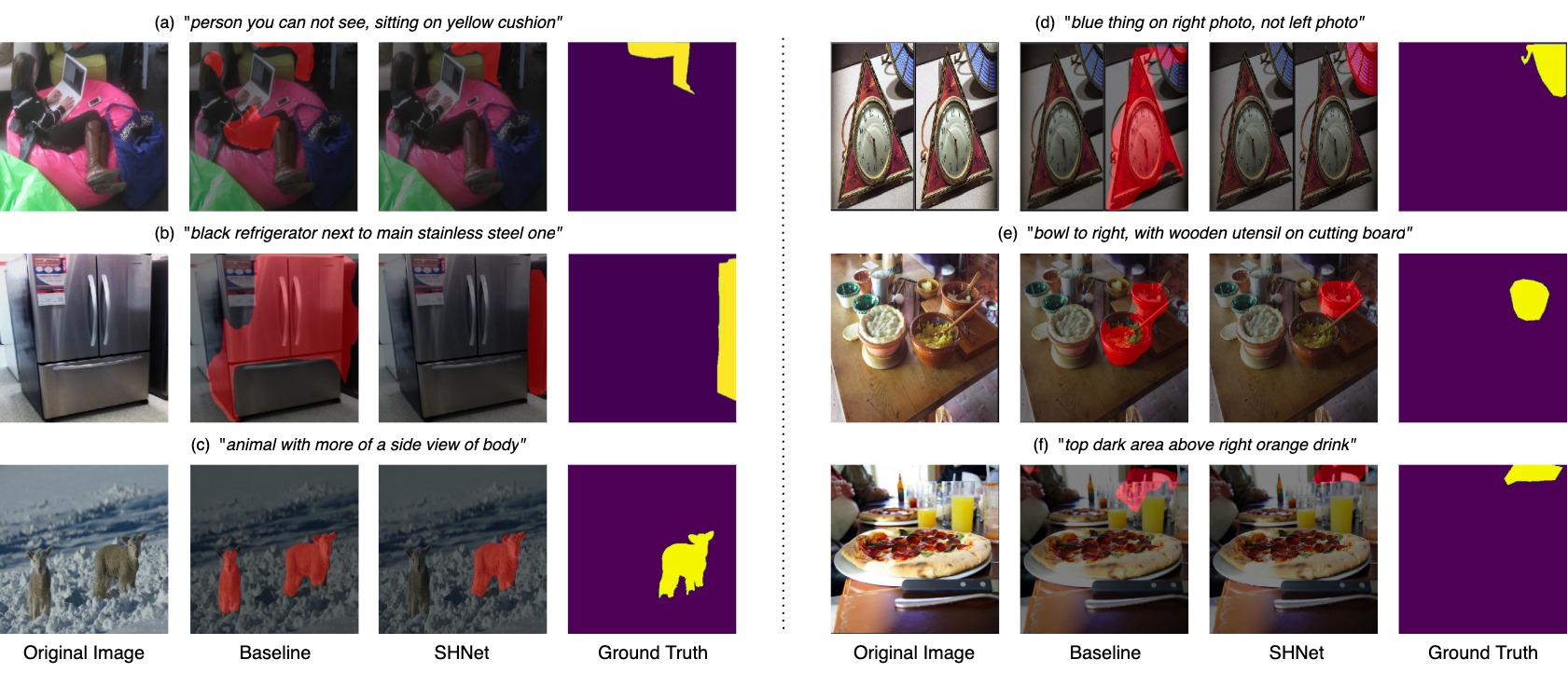}
    \caption{Qualitative results comparing the baseline against SHNet.}
    \label{fig:qualitative}
\end{figure*}


\subsection{Ablation Studies}

We perform ablation studies on the UNC dataset's validation split. All methods are evaluated on Precision@$X$ and Overall IoU metrics, and the results are illustrated in Table~\ref{table:ablation_1}. Unless specified, the backbone used for ablations is DeepLabv3+ trained at $320 \times 320$ resolution. The feature extraction process described in Section 3.1 is used for all ablation studies. ASPP + ConvUpsample decoder is also common to all the experiments. 


\emph{\textbf{Baseline}}: The baseline model involves direct concatenation of visual features with the tiled textual feature to result in multi-modal feature of shape $\mathbb R^{(C_v + C_l) \times H \times W}$. This multi-modal feature is passed as input to ASPP + ConvUpsample decoder.

\begin{figure*}[t]
\begin{center}
\begin{tabular}{c c c c c} 
\multicolumn{5}{c}{\centering{\textit{\scriptsize{}``the right half of the sandwich on the left"}}}\\
        \includegraphics[width=0.13\linewidth]{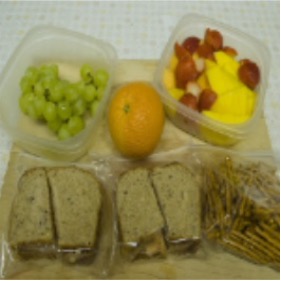}& 
        \includegraphics[width=0.13\linewidth]{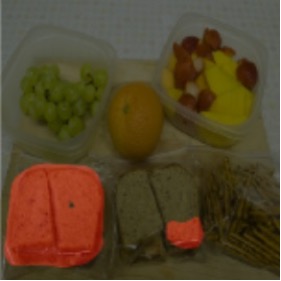}&
        \includegraphics[width=0.13\linewidth]{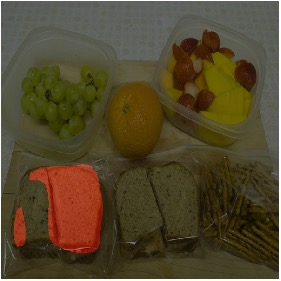}& 
        \includegraphics[width=0.13\linewidth]{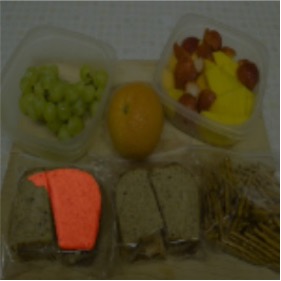}&
        \includegraphics[width=0.13\linewidth]{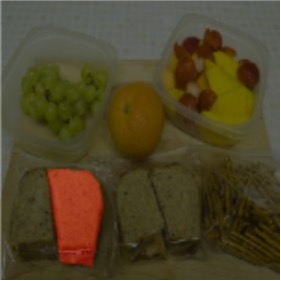}\\
         \centering{\scriptsize{(a) Original Image}} &
         \centering{\scriptsize{(b) Only HCAM module}} &
         \centering{\scriptsize{(c) Only SFM module}} &
         \centering{\scriptsize{(d) SHNet}} &
         \centering{\scriptsize{(e) Ground Truth}} \vspace{-1em}
         \end{tabular}
\end{center} 
    \caption{ Qualitative results corresponding to combinations of proposed modules. In (b) we show results when only HCAM module is used, (c) result with only SFM module being used, (d) output mask when both SFM and HCAM modules are used}
    \label{fig:Module_Ablation}
\end{figure*}

\begin{figure*}[t]
    \centering
    \begin{tabular}[h]{p{0.14\textwidth} p{0.14\textwidth} p{0.14\textwidth}                            p{0.14\textwidth} p{0.14\textwidth} p{0.145\textwidth}}
        \includegraphics[width=\linewidth]{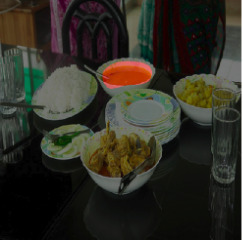}& 
        \includegraphics[width=\linewidth]{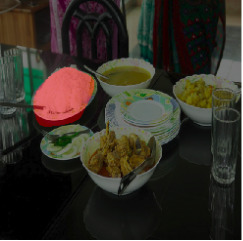}&
        \includegraphics[width=\linewidth]{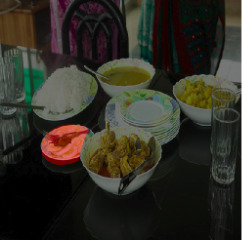}&
        \includegraphics[width=\linewidth]{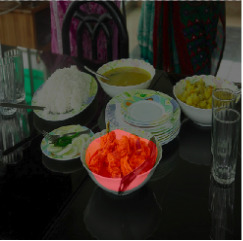}&
        \includegraphics[width=\linewidth]{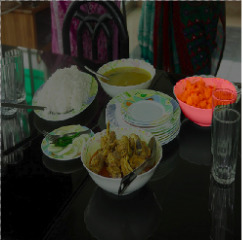}&
        \includegraphics[width=0.98\linewidth]{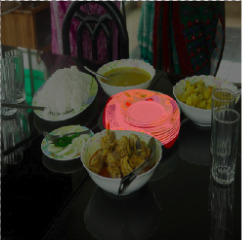}\\
        \centering{\scriptsize{``top bowl"}}&  
        \centering{\scriptsize{``left plate on top"}}&
        \centering{\scriptsize{``left plate on bottom"}}&
        \centering{\scriptsize{``front bowl"}}&  
        \centering{\scriptsize{``right bowl"}}&  
        \centering{\scriptsize{``empty plates in center"}} 
    \end{tabular}
    \caption{Output predictions of SHNet for an anchored image with varying linguistic expressions.}
    \label{fig:pivot}
\end{figure*}


\emph{\textbf{HCAM without SFM}}: ``Only HCAM" network differs with baseline method only on the fusion process of hierarchical multi-modal features. Introducing the HCAM module over baseline results in 4.83 \% improvement on the Overall IoU metric and an improvement of 2.5 \% on the $ prec@0.9 $ metric (illustrated in Table~\ref{table:ablation_1}), indicating that the HCAM module results in refined segmentation masks.

\emph{\textbf{SFM without HCAM}}: Similarly, the ``Only SFM" network differs from the baseline method in how different types of visual-linguistic interactions are captured. We observe significant performance gains of 7.46 \% over the baseline, indicating that simultaneous interactions help identify the referent.


\emph{\textbf{SFM + X}}: We replace HCAM module with other multi-level fusion techniques like ConvLSTM and Conv3D. Comparing the performance of SFM+ConvLSTM with SHNet (SFM+HCAM), we observe that HCAM is indeed effective at fusing hierarchical multi-modal features (Table~\ref{table:ablation_1}). For SFM+Conv3D, we stack multi-level features along a new depth dimension resulting in 3D features, and perform 3D convolution on them. The same filter is applied to different level features that result in each level feature converging on a common region in the image. SFM+Conv3D achieves a similar performance as SFM+ConvLSTM while using fewer parameters. Using Conv3D achieves higher Precision@$0.8$ and Precision@$0.9$ than ConvLSTM, suggesting that it leads to more refined maps. It is worth noting that HCAM also uses Conv3D at the end, and the additional gains of SHNet over SFM+Conv3D suggest the benefits of hierarchical information exchange in HCAM.  

\emph{\textbf{Glove and Positional Embeddings}}: We verify Glove embeddings' significance by replacing it with one hot embedding. We also validate the usefulness of Positional Embeddings (P.E.) by training a model without them. Both variants observe a drop in performance (Table~\ref{table:ablation_1}), with the drop being more significant in the variant without Glove embeddings. These ablations suggest the importance of capturing word-level semantics and positional-aware features.




In Table~\ref{table:resolution_backbone}, we present ablations with different backbones at different resolution.
The results demonstrate that our approach does not heavily rely on backbone for its performance gains, as even with a vanilla Imagenet pre-trained Resnet101 backbone, not fine-tuned on segmentation task, we outperform existing methods at both resolutions. Predictably, using a backbone fine-tuned on a segmentation task gives further performance gain.

\begin{table}[h]
\small
\begin{center}
\begin{tabular}{|c|c|c|c|c|}
\hline
backbone                         & resolution & val   & testA & testB \\ \hline
\multirow{2}{*}{Resnet101}       & 320 x 320  & 63.76 & 67.05 & 60.15 \\ \cline{2-5} 
                                 & 448 x 448  & 64.88 & 68.08 & 60.82 \\ \hline
\multirow{2}{*}{DeepLabv3+}      & 320 x 320  & 63.98 & 67.51 & 60.48 \\ \cline{2-5} 
                                 & 448 x 448  & 65.29 & 68.56 & 62.04 \\ \hline
\end{tabular}
\end{center}
\caption{Result with different backbone at different input resolutions on UNC dataset.}
\label{table:resolution_backbone}
\end{table}

\begin{table}[h]
\small
\begin{center}
\begin{tabular}{|c|c|l|}
\hline
\multirow{2}{*}{
    \begin{tabular}[c]{@{}c@{}}
    Aggregation Module
    \end{tabular}
}
                        & \multicolumn{2}{c|}{{ Overall IOU}}                     \\ \cline{2-3} 
                        & \multicolumn{1}{l|}{{ 320x320}} & {{448x448}}           \\ \hline
{MGATE~\cite{ye2019cross}}        & {\centering{62.59}}  & {\centering{63.35}}    \\ \hline
{TGFE~\cite{Huang_2020_CVPR}}     & {\centering{62.94}}  & {\centering{63.72}}    \\ \hline
{GBFM~\cite{hui2020linguistic}}   & {\centering{62.72}}  & {\centering{63.83}}    \\ \hline
{HCAM}                           & {\centering{\textbf{63.98}}}  & {\centering{\textbf{65.32}}}  \\ \hline
\end{tabular}
\end{center}
\caption{Comparing performance of recent Aggregation Modules on the UNC val dataset at different input resolutions}
\label{table:fusion_comparison}
\end{table}

We also present ablations with different aggregation modules in Table~\ref{table:fusion_comparison}. We use the modules presented in MGATE~\cite{ye2019cross}, TGFE~\cite{Huang_2020_CVPR} and GBFM~\cite{hui2020linguistic}, for which codes were publicly available. HCAM consistently outperforms other methods by clear margins at both resolution.

\subsection{Qualitative Results}

\begin{figure}[t]
\begin{center}
  \includegraphics[width=\linewidth]{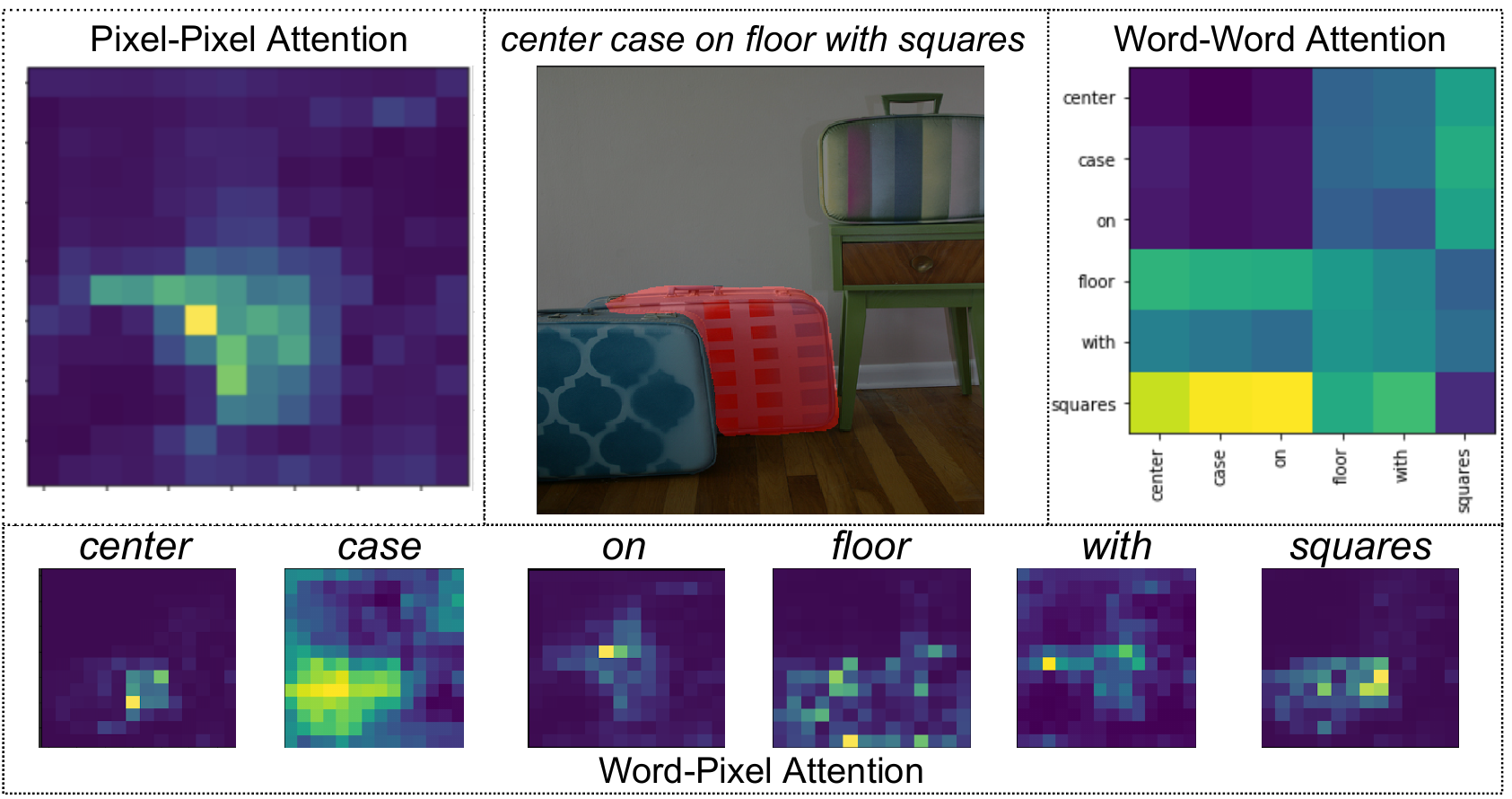}
\end{center}
  \caption{Visualization of Inter-modal and Intra-modal interactions in SFM.}
\label{fig:attention_vis}
\end{figure}

Figure \ref{fig:qualitative} presents qualitative results comparing SHNet against the baseline model. SHNet localizes heavily occluded objects (Figure \ref{fig:qualitative} (a) and (b)); reasons on the overall essence of the highly ambiguous sentences (e.g. ``person you cannot see", ``right photo not left photo") and; distinguishes among multiple instances of the same type of object based on attributes and appearance cues (Figure \ref{fig:qualitative} (b), (c), and (e)). While, without any reasoning stage, the baseline model struggles to segment the correct instance and confuse it with similar objects. Figure \ref{fig:qualitative} (d) and (f) illustrate the ability of SHNet to localize unstructured non-explicit objects like ``dark area" and ``blue thing". The potential of SHNet to perform relative positional reasoning is highlighted in Figure \ref{fig:qualitative} (b), (e), and (f).

We outline the contributions of both SFM and HCAM modules in Figure~\ref{fig:Module_Ablation}. ``Only HCAM" network does not involve any reasoning, however, it manages to predict the left sandwich with refined boundaries. ``Only SFM" network understands the concept of ``the right half of the sandwich" and leads to much better output; however, the output mask bleeds around the boundaries, and an extra small noisy segment is visible. The full model benefits from the reasoning in ``SFM," and when combined with HCAM facilitates information exchange across hierarchies to predict correct refined mask as output. In Figure~\ref{fig:pivot}, we anchor an image and vary the linguistic expression. SHNet is able to reason about different linguistic expressions successfully and ground them. Inter-modal and Intra-modal interactions captured by SFM are illustrated in Figure \ref{fig:attention_vis}. Pixel-pixel interactions highlight image regions corresponding to the referent. For the given expression, ``squares" contains the differentiating information and is assigned high importance for different words. Additionally, for each word appropriate region in the image is attended.




\section{Conclusion}
In this work, we tackled the task of Referring Image Segmentation. We proposed a simple yet effective SFM to capture comprehensive interactions between modalities in a single step, allowing us to simultaneously consider the contextual information from both modalities. Furthermore, we introduced a novel HCAM module to aggregate multi-modal context across hierarchies. Our approach achieves strong performance on RIS benchmarks without any post-processing. We present thorough quantitative and qualitative experiments to demonstrate the efficacy of all the proposed components.

\bibliography{main}
\bibliographystyle{main}

\end{document}